\documentstyle[acl93,epsf]{article}
\newcommand{\beq}[1]{\begin{equation}\label{#1}}
\newcommand{\eeq}{\end{equation}}
\newcommand{\bseq}[1]{\begin{displaymath}}
\newcommand{\eseq}{\end{displaymath}}
\newcommand{\beqa}[1]{\begin{equation}\label{#1}\begin{eqalign}}
\newcommand{\eeqa}{\end{eqalign}\end{equation}}
\newcommand{\bsubeq}[1]{\begin{subequations}\label{#1}\begin{eqalignno}}
\newcommand{\esubeq}{\end{eqalignno}\end{subequations}}

\author{Fernando Pereira \\
AT\&T Bell Laboratories \\ 600 Mountain
Ave. \\ Murray Hill, NJ 07974 \\{\tt pereira@research.att.com}\And
Naftali Tishby \\
Dept.~of Computer
Science \\  Hebrew University \\ Jerusalem 91904, Israel \\
{\tt tishby@cs.huji.ac.il}\And
Lillian Lee \\
Dept.~of Computer Science \\ Cornell
University \\ Ithaca, NY \\
{\tt llee@cs.cornell.edu}}
\title{DISTRIBUTIONAL CLUSTERING OF ENGLISH WORDS}
\begin{document}
\maketitle
\begin{abstract}
We describe and experimentally evaluate a method for automatically
clustering words according to their distribution in particular
syntactic contexts. Deterministic annealing is used to find lowest
distortion sets of clusters. As the annealing parameter increases,
existing clusters become unstable and subdivide, yielding a
hierarchical ``soft'' clustering of the data. Clusters are used as the
basis for class models of word coocurrence, and the models evaluated
with respect to held-out test data.
\end{abstract}
\section{INTRODUCTION}
Methods for automatically classifying words according to their
contexts of use have both scientific and practical interest. The
scientific questions arise in connection to distributional views of
linguistic (particularly lexical) structure and also in relation to
the question of lexical acquisition both from psychological and
computational learning perspectives. From the practical point of view,
word classification addresses questions of data sparseness and
generalization in statistical language models, particularly
models for deciding among alternative analyses proposed by a grammar.

It is well known that a simple tabulation of frequencies of certain
words participating in certain configurations, for example of
frequencies of pairs of a transitive main verb and the head noun of
its direct object, cannot be reliably used for comparing the
likelihoods of different alternative configurations. The problem is
that for large enough corpora the number of possible joint events is
much larger than the number of event occurrences in the corpus, so
many events are seen rarely or never, making their frequency counts
unreliable estimates of their probabilities.

Hindle \shortcite{Hindle:ACL90} proposed dealing with the sparseness
problem by estimating the likelihood of unseen events from that of
``similar'' events that have been seen. For instance, one may estimate
the likelihood of a particular direct object for a verb from the
likelihoods of that direct object for similar verbs. This
requires a reasonable definition of verb similarity and a similarity
estimation method.  In Hindle's proposal, words are similar if we have
strong statistical evidence that they tend to participate in the same
events. His notion of similarity seems to agree with our intuitions
in many cases, but it is not clear how it can
be used directly to construct word classes and corresponding models of
association.

Our research addresses some of the same questions and uses similar
raw data, but we investigate how to factor word association
tendencies into associations of words to certain hidden {\em
senses classes} and associations between the classes themselves. While
it may be worthwhile to base such a model on preexisting sense classes
\cite{Resnik:WordNet}, in the work described here we look at how
to derive the classes directly from distributional data. More
specifically, we model senses as probabilistic concepts or {\em
clusters} $c$ with
corresponding cluster membership probabilities $p(c|w)$ for each
word $w$. Most other class-based modeling techniques for natural
language rely instead on ``hard'' Boolean classes
\cite{Brown+al:class}. Class construction is then combinatorially very
demanding and depends on frequency counts for joint events involving
particular words, a potentially unreliable source of information as we
noted above. Our approach avoids both problems.

\subsection{Problem Setting}
\label{problem}
In what follows, we will consider two major word classes, ${\cal V}$
and ${\cal N}$, for the verbs and nouns in our experiments, and a
single relation between them, in our experiments relation between a
transitive main verb and the head noun of its direct object. Our raw
knowledge about the relation consists of the frequencies $f_{vn}$ of
occurrence of particular pairs $(v,n)$ in the required configuration
in a training corpus. Some form of text analysis is required to
collect such a collection of pairs. The corpus used in our first
experiment was derived from newswire text automatically parsed by
Hindle's parser Fidditch \cite{Hindle:fidd}. More recently, we have
constructed similar tables with the help of a statistical
part-of-speech tagger
\cite{Church:parts} and of tools for regular expression pattern
matching on tagged corpora \cite{Yarowsky:conc}. We have not yet
compared the accuracy and coverage of the two methods, or what
systematic biases they might introduce, although we took care to filter
out certain systematic errors, for instance the misparsing of the
subject of a complement clause as the direct object of a main verb for
report verbs like ``say''.

We will consider here only the problem of classifying nouns according
to their distribution as direct objects of verbs; the converse problem
is formally similar. More generally, the theoretical basis for our
method supports the use of clustering to build models for any $n$-ary
relation in terms of associations between elements in each coordinate
and appropriate hidden units (cluster centroids) and associations
between those hidden units.

For the noun classification problem, the empirical distribution of a
noun $n$ is then given by the conditional density
$p_n(v)=f_{vn}/\sum_v f_{vn}$. The problem we study is how to use the
$p_n$ to classify the $n\in {\cal N}$. Our classification method will
construct a set ${\cal C}$ of clusters and cluster membership
probabilities $p(c|n)$. Each cluster $c$ is associated to a cluster
{\em centroid} $p_c$, which is discrete density over ${\cal V}$
obtained by averaging appropriately the $p_n$.

\subsection{Distributional Similarity}

To cluster nouns $n$ according to their conditional verb distributions
$p_n$, we need a measure of similarity between distributions. We use
for this purpose the {\em relative entropy} or {\em Kullback-Leibler (KL)
distance} between two distributions
\[
D(p\parallel q) = \sum_x p(x) \log \frac{p(x)}{q(x)}{\rm\quad .}
\]
This is a natural choice for a variety of reasons, which we will just
sketch here.\footnote{A more formal discussion will appear in our paper
{\em Distributional Clustering}, in preparation.}

First of all, $D(p\parallel q)$ is zero just in case $p=q$, and it
increases as the probability decreases that $p$ is the relative
frequency distribution of a random sample drawn according to $p$. More
formally, the probability mass given by $q$ to the set of all samples
of length $n$ with relative frequency distribution $p$ is bounded by
$2^{-nD(p\parallel q)}$ \cite{Cover+Thomas:info}. Therefore, if we are
trying to distinguish among hypotheses $q_i$ when $p$ is the relative
frequency distribution of observations, $D(p\parallel q_i)$ gives the
relative weight of evidence in favor of $q_i$. Furthermore, a similar
relation holds between $D(p \parallel p')$ for two empirical
distributions $p$ and $p'$ and the probability that $p$ and $p'$ are
drawn from the same distribution $q$. We can thus use the relative
entropy between the context distributions for two words to measure how
likely they are to be instances of the same cluster centroid.

{}From an information theoretic perspective $D(p\parallel q)$
measures how inefficient on average it would be to use a code based on
$q$ to encode a variable distributed according to $p$. With respect to
our problem, $D(p_n\parallel p_c)$ thus gives us the loss of
information in using cluster centroid $p_c$ instead of the actual
distribution for word $p_n$ when modeling the distributional
properties of $n$.

Finally, relative entropy is a natural measure of similarity between
distributions for clustering because its minimization leads to cluster
centroids that are a simple weighted average of member distributions.

One technical difficulty is that $D(p\parallel p')$ is not defined
when $p'(x) = 0$ but $p(x) > 0$. We could sidestep this problem (as we
did initially) by smoothing zero frequencies appropriately
\cite{Church+Gale:GT}.  However, this is not very satisfactory because
one of the goals of our work is precisely to avoid the problems of
data sparseness by grouping words into classes. It turns out that the
problem is avoided by our clustering technique, since it does not need
to compute the KL distance between individual word distributions, but
only between a word distribution and average distributions, the
current cluster centroids, which are guaranteed to be nonzero whenever
the word distributions are.  This is a useful advantage of our method
compared with agglomerative clustering techniques that need to compare
individual objects being considered for grouping.

\section{THEORETICAL BASIS}
\label{theory}
In general, we are interested on how to organize a set of linguistic
objects such as words according to the contexts in which they occur,
for instance grammatical constructions or $n$-grams.  We will show
elsewhere that the theoretical analysis outlined here applies to that
more general problem, but for now we will only address the more
specific problem in which the objects are nouns and the contexts are
verbs that take the nouns as direct objects.

Our problem can be seen as that of learning a joint distribution of
pairs from a large sample of pairs. The pair coordinates come from two
large sets ${\cal N}$ and ${\cal V}$, with no preexisting topological
or metric structure, and the training data is a sequence $S$ of $N$
independently drawn pairs
\bseq{sentence}
S_i = ( n_{i} , v_{i} ) \qquad 1\leq i \leq N\;.
\eseq
{}From a learning perspective, this problem falls somewhere in between
unsupervised and supervised learning. As in unsupervised learning, the
goal is to learn the underlying distribution of the data. But in contrast
to most unsupervised learning settings, the objects involved
have no internal structure or attributes allowing them to be compared
with each other. Instead, the only information about the objects is
the statistics of their joint appearance. These statistics can thus be
seem as a weak form of object labelling analogous to supervision.

\subsection{Distributional Clustering}
While clusters based on distributional similarity are interesting on
their own, they can also be profitably seen as a means of summarizing
a joint distribution. In particular, we would like to find a set of
clusters ${\cal C}$ such that each conditional
distribution $p_n(v)$ can be approximately decomposed as
\bseq{decom2}
\hat{p}_n (v) =
   \sum_{c \in \cal C} p(c | n ) p_c(v) \quad,
\eseq
where $ p(c | n )$ is the membership probability of $n$ in $c$ and $
p_c(v) = p(v|c)$ is $v$'s conditional probability given by the
centroid distribution for cluster $c$.

The above decomposition can be written in a more symmetric form as
\begin{eqnarray}
\hat{p}(n,v) & = &\sum_{c \in \cal C} p(c,n) p(v|c) \nonumber \\
& = & \sum_{c \in \cal C} p(c) p(n|c) p(v|c) \label{decom3}
\end{eqnarray}
assuming that $p(n)$ and $\hat{p}(n)$ coincide.
We will take (\ref{decom3}) as our basic clustering model.

To determine this decomposition we need to solve the two connected
problems of finding find suitable forms for the cluster membership and
centroid distributions $p(v|c)$, and of maximizing the goodness of fit
between the model distribution $\hat{p}(n,v)$ and the observed data

Goodness of fit is determined by the model's likelihood of the
observations. The maximum likelihood (ML) estimation principle is thus
the natural tool to determine the centroid distributions $p_c(v)$.

As for the membership probabilities, they must be determined solely by
the relevant measure of object-to-cluster similarity, which in the
present work is the relative entropy between object and cluster
centroid distributions.  Since no other information is available, the
membership is determined by maximizing the configuration entropy
subject for a fixed average distortion.  With the
maximum entropy (ME) membership distribution, ML estimation is
equivalent to the minimization of the average distortion of the data.
The combined entropy maximization entropy and distortion minimization
is carried out by a two-stage iterative process similar to the
EM method
\cite{Dempster+al:EM}. The first stage of an iteration is a maximum
likelihood, or minimum distortion, estimation of the cluster centroids
given fixed membership probabilities.  In the second iteration stage,
the entropy of the membership distribution is maximized with a fixed
average distortion.  This joint optimization searches for a {\em
saddle point} in the distortion-entropy parameters, which is
equivalent to minimizing a linear combination of the two
known as {\em free energy} in statistical mechanics. This analogy with
statistical mechanics is not coincidental, and provide us with a
better understanding of the clustering procedure.

\subsubsection{Maximum Likelihood Cluster Centroids}
For the maximum likelihood argument, we start by estimating the
likelihood of the sequence $S$ of $N$ independent observations of
pairs $(n_i , v_i )$.  Using
(\ref{decom3}), the sequence's model log likelihood is
\bseq{log_likelihood1}
l(S) = \log \hat{p}(S)
= \sum_{i=1}^{N} \log \sum_{c \in {\cal C}} p(c) p( n_i | c) p( v_i |c)\quad.
\eseq
Fixing the number of clusters (model size) $|\cal C |$, we want to
maximize $l(S)$ with respect to the distributions
$p(n | c )$ and $p(v|c)$.  The variation of $l(S)$   with
respect to these distributions is
\beq{like_var}
\delta { l} (S) = \sum_{i=1}^{N} \frac{1}{\hat{p}(n_i , v_i )}
 \sum_{c \in {\cal C}} p(c) \left( \begin{array}{c} p(v_i | c)
\delta p( n_i | c) \\ + \\
p(n_i | c) \delta p( v_i |c )\end{array} \right)
\eeq
with $p(n|c)$ and $p(v|c)$ kept normalized.  Using Bayes's
formula, we have \footnote{As usual in clustering models
\cite{Duda+Hart:class}, we assume that the model distribution and the
empirical distribution are interchangeable at the solution of the
parameter estimation equations, since the model is assumed to be able
to represent correctly the data at that solution point. In practice,
the data may not come exactly from the chosen model class, but
the model obtained by solving the
estimation equations may still be the closest one to the data.}
\[
p(n_i | c)p(v_i |c) = \frac{p( c | n_i , v_i )}{p(c)} \hat{p}(n_i , v_i ) ~,
\]
 or
\[
\frac{1}{ \hat{p}(n_i , v_i ) } = \frac{p( c | n_i ,v_i )}{p(c) p(n_i | c)
p(v_i |c )}
\]
for any $c$, which we substitute into (\ref{like_var}) to obtain
\beq{like_var1}
\delta { l} (S) =   \sum_{i=1}^{N}
 \sum_{c \in {\cal C}} p(c | n_i , v_i )
\left(\begin{array}{c}\delta \log p( n_i | c) \\ + \\
 \delta \log p(v_i |c)\end{array}\right)
\eeq
since $\delta \log p = \delta p / p$.
This expression is particularly useful when the cluster distributions
$p(n|c)$ and $p(v|c)$ are of exponential form,
precisely what will be provided by the ME step described below.

At this point we need to specify the clustering model in more detail.
In the derivation so far we have treated $p(n|c)$ and $p(v|c)$
symmetrically, corresponding to clusters not of verbs or nouns but of
verb-noun associations. In principle such a symmetric model may be
more accurate, but in this paper we will concentrate on {\em
asymmetric models} in which cluster memberships are associated to just
one of the components of the joint distribution and the cluster
centroids are specified only by the other component.  In particular,
the model we use in our experiments has noun clusters with cluster
memberships determined by $p(n|c)$ and centroid distributions
determined by $p(v|c)$.

The asymmetric model simplifies the estimation significantly by
dealing with a single component, but it has the disadvantage that the
joint distribution, $p(n,v)$ has two different and not necessarily
consistent expressions in terms of asymmetric models for the two
coordinates.
\subsubsection{Maximum Entropy  Cluster Membership}
While variations of $ p(n|c)$ and $p(v|c)$ in equation
(\ref{like_var1} are not independent, we can treat them separately.
First, for fixed average distortion between the cluster centroid
distributions $p(v|c)$ and the data $p(v|n)$, we find the
cluster membership probabilities, which are the Bayes's inverses of
the  $ p(n|c)$, that maximize the entropy of the cluster distributions.
With the membership distributions thus obtained, we then look
for the $p(v|c)$ that maximize the log likelihood $l(S)$. It turns out
that this will also be the values of $p(v|c)$ that minimize the
average distortion between the asymmetric cluster model and the data.

Given any similarity measure $d(n,c)$ between nouns and cluster
centroids, the
average cluster distortion is
\beq{average_dist}
\langle D \rangle = \sum_{n \in {\cal N}}  \sum_{c\in {\cal C}}
p( c | n ) d(n,c)
\eeq
If we maximize the cluster membership entropy
\beq{class_ent}
H = - \sum_{n \in {\cal N}}\sum_{c\in {\cal C}} p(c|n) \log p(n|c)
\eeq
subject to normalization of $p(n|c)$ and fixed (\ref{average_dist}), we obtain
the following standard exponential forms for the class and membership
distributions
\beq{exp_form_c}
p(n |c ) = \frac{1}{Z_c} \exp  - \beta d(n,c)
\eeq
\beq{exp_form_x}
p(c |n ) = \frac{1}{Z_n} \exp  - \beta d(n,c)
\eeq
where the normalization sums (partition functions) are $Z_c = \sum_n
\exp - \beta d(n,c)$ and $Z_n = \sum_c \exp - \beta d(n,c)$. Notice
that $d(n,c)$ does not need to be symmetric for this derivation, as
the two distributions are simply related by Bayes's rule.

Returning to the log-likelihood variation (\ref{like_var1}),
we can now use (\ref{exp_form_c}) for $p(n|c)$ and the
assumption for the asymmetric model that the cluster membership
stays fixed as we adjust the centroids, to obtain
\beq{like_var2}
\delta { l} (S) = -  \sum_{i=1}^{N}
 \sum_{c \in {\cal C}} p(c | n_i ) \delta \beta d(n_i,c)
+ \delta \log Z_c
\eeq
where the variation of $p(v|c)$ is now included in the variation of
$d(n,c)$.

For a large enough sample, we may replace the sum over observations in
(\ref{like_var2}) by the average over ${\cal N}$
\bseq{like_var3}
\delta {l} (S) = -  \sum_{n\in N}
 p(n) \sum_{c \in {\cal C}} p(c | n ) \delta \beta d(n,c)
+ \delta \log Z_c
\eseq
which, applying Bayes's rule, becomes
\beq{like_var4}
\delta {l} (S) = -  \sum_{c \in {\cal C}}
 \frac{1}{p(c)} \sum_{n\in N} p(n | c ) \delta \beta d(n,c)
+ \delta \log Z_c
\eeq
At the log-likelihood maximum, the
variation (\ref{like_var4}) must vanish. We will see below
that the use of relative entropy for similarity measure makes
$\delta \log Z_c$ vanish at the maximum as well, so the
log likelihood can be maximized by minimizing the average distortion
with respect to the class centroids while class
membership is kept fixed
\[
\sum_{c\in{\cal C}} \frac{1}{p(c)} \sum_{n\in{\cal  N}}  p( n |c ) \delta d(n,
c) = 0\quad,
\]
or, sufficiently, if each of the inner sums vanish
\beq{min_dist}
\sum_{c\in{\cal C}} \sum_{n\in {\cal N}}  p( n |c ) \delta d(n, c) = 0
\eeq

\subsubsection{Minimizing the Average KL Distortion}
We first show that the minimization of the relative entropy yields the
natural expression for cluster centroids
\begin{equation}
p(v|c) = \sum_{n\in {\cal N}}p(n|c)p(v|n) \label{centroid}
\end{equation}
To minimize the
average distortion (\ref{min_dist}), we observe that the variation of
the KL distance between noun and centroid distributions with respect
to the centroid distribution $p(v|c)$, with
each centroid distribution normalized by the Lagrange multiplier
$\lambda_c$, is given by
\begin{eqnarray*}
\delta d(n,c) &=& \delta \left( \begin{array}{c}
 -\sum_{v\in {\cal V}} p(v|n ) \log p(v|c) \\
+ \\ \lambda_c ( \sum_{v\in {\cal V}} p(v|c) -1) \end{array} \right)\\
&=& \sum_{v \in {\cal V}} \left(-\frac{p(v|n ) }{p(v|c)}
+ \lambda_c\right)
\delta p(v|c) \quad.
\end{eqnarray*}
Substituting this expression into (\ref{min_dist}), we obtain
\bseq{delD1}
\sum_c \sum_n \sum_v \left( -\frac{p(v|n ) p(n|c )}{p(v|c)}
+ \lambda_c \right) \delta p(v|c) =0\quad.
\eseq
Since the $\delta p(v|c)$
are now independent, we obtain immediately the desired centroid
expression (\ref{centroid}), which is the desired
weighted average of noun distributions.

We can now see that the variation $\delta \log Z_c$ vanishes for
centroid distributions given by  (\ref{centroid}), since it follows
{}from (\ref{min_dist}) that
\begin{eqnarray*}
\delta \log Z_c & = &-\frac{\beta}{Z_c}\sum_n \exp -\beta d(n,c)
\delta d(n,c) \\ & = &-\beta \sum_n p(n|c) \delta d(x,c) = 0.
\end{eqnarray*}

\subsubsection{The Free Energy Function}
The combined minimum distortion and maximum entropy optimization is
equivalent to the minimization of a single function,
the {\em free energy}
\begin{eqnarray*}
F& = &-\frac{1}{\beta} \sum_n \log Z_n\\
& = & \langle D\rangle - H/\beta
\end{eqnarray*}
where $\langle D\rangle$ is the average distortion
(\ref{average_dist}) and $H$ is the cluster membership entropy
(\ref{class_ent}).

The free energy determines both the distortion and the membership entropy
through
\begin{eqnarray*}
\langle D\rangle &=& \frac{\partial \beta F}{\partial \beta}\\
  H &=& - \frac{\partial F}{\partial T} \quad,
\end{eqnarray*}
with {\em temperature} $T = \beta^{-1}$.

The most important property of the free energy is that its minimum
determines the balance between the ``disordering'' maximum entropy and
``ordering'' distortion minimization in which the system is most
likely to be found. In fact
the probability to find the system at a given configuration is
exponential in $F$
\bseq{likefree}
P \propto \exp {- \beta F} \quad,
\eseq
so a system is most likely to be found in its minimal free energy
configuration.

\subsection{Hierarchical Clustering}

The analogy with statistical mechanics suggests a {\em deterministic
annealing} procedure for clustering \cite{Rose+al:phase}, in which
the number of clusters is determined through a sequence of phase
transitions by continuously increasing the parameter $\beta$
following an {\em annealing schedule}.

The higher $\beta$, the more local is the
influence of each noun on the definition of centroids.  The
dissimilarity plays here the role of distortion.  When the scale
parameter $\beta$ is close to zero, the dissimilarities are almost
irrelevant, all words contribute about equally to each centroid, and
so the lowest average distortion solution involves just one cluster
which is the average of all word densities. As $\beta$ is slowly
increased, a point (phase transition) is eventually reached which the
natural solution involves two distinct centroids. We say then that the
original cluster has {\em split} into the two new clusters.

In general, if we take any cluster $c$ and a {\em twin} $c'$ of $c$
such that the centroid $p_{c'}$ is a small random pertubation of
$p_c$, below the critical $\beta$ at which $c$ splits the membership
and centroid reestimation procedure given by equations
(\ref{exp_form_x}) and (\ref{centroid}) will make $p_c$ and $p_{c'}$
converge, that is, $c$ and $c'$ are really the same cluster. But with
$\beta$ above the critical value for $c$, the two centroids will
diverge, giving rise to two daughters of $c$.

Our clustering procedure is thus as follows.  We start with very low
$\beta$ and a single cluster whose centroid is the average of all noun
distributions. For any given $\beta$, we have a current set of {\em
leaf} clusters corresponding to the current free energy (local)
minimum. To refine such a solution, we search for the lowest $\beta$
which is the critical value for some current leaf cluster splits.
Ideally, there is just one split at that critical value, but
for practical performance and numerical accuracy reasons we may have
several splits at the new critical point. The splitting procedure can
then be repeated to achieve the desired number of clusters or model
cross-entropy.
\begin{figure}
\setlength{\epsfxsize}{3in}
\centerline{\mbox{\epsffile{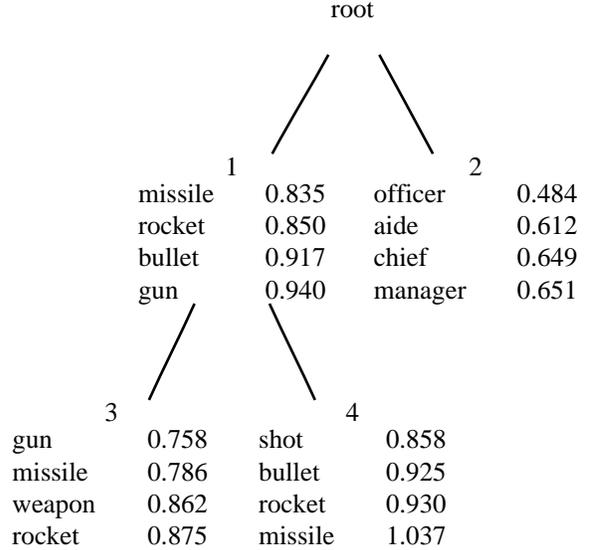}}}
\caption{Direct object clusters for {\em fire}}
\label{fire-clusters}
\end{figure}
\section{CLUSTERING EXAMPLES}
\label{sec:examples}
All our experiments involve the asymmetric model described in the
previous section. As explained there, our clustering procedure yields
for each value of $\beta$ a set $C_\beta$ of clusters minimizing the
free energy $F$, and the asymmetric model for $\beta$ estimates the
conditional verb distribution for a noun $n$ by
\[ \hat{p}_n = \sum_{c\in C_\beta} p(c|n) p_c \]
where $p(c|n)$ also depends on $\beta$.

As a first experiment, we used our method to classify the 64 nouns
appearing most frequently as heads of direct objects of the verb
``fire'' in one year (1988) of Associated Press newswire. In this
corpus, the chosen nouns appear as direct object heads of a total of
2147 distinct verbs, so each noun is represented by a density over
the 2147 verbs.

Figure \ref{fire-clusters} shows the five words most similar to the each
cluster centroid for the four clusters resulting from the first two
cluster splits.  It can be seen that first split separates the objects
corresponding to the weaponry sense of ``fire'' (cluster 1) from the
ones corresponding to the personnel action (cluster 2). The second
split then further refines the weaponry sense into a projectile sense
(cluster 3) and a gun sense (cluster 4). That split is
somewhat less sharp, possibly because not enough distinguishing
contexts occur in the corpus.

\begin{figure*}
\setlength{\epsfxsize}{5.5in}
\centerline{\mbox{\epsffile{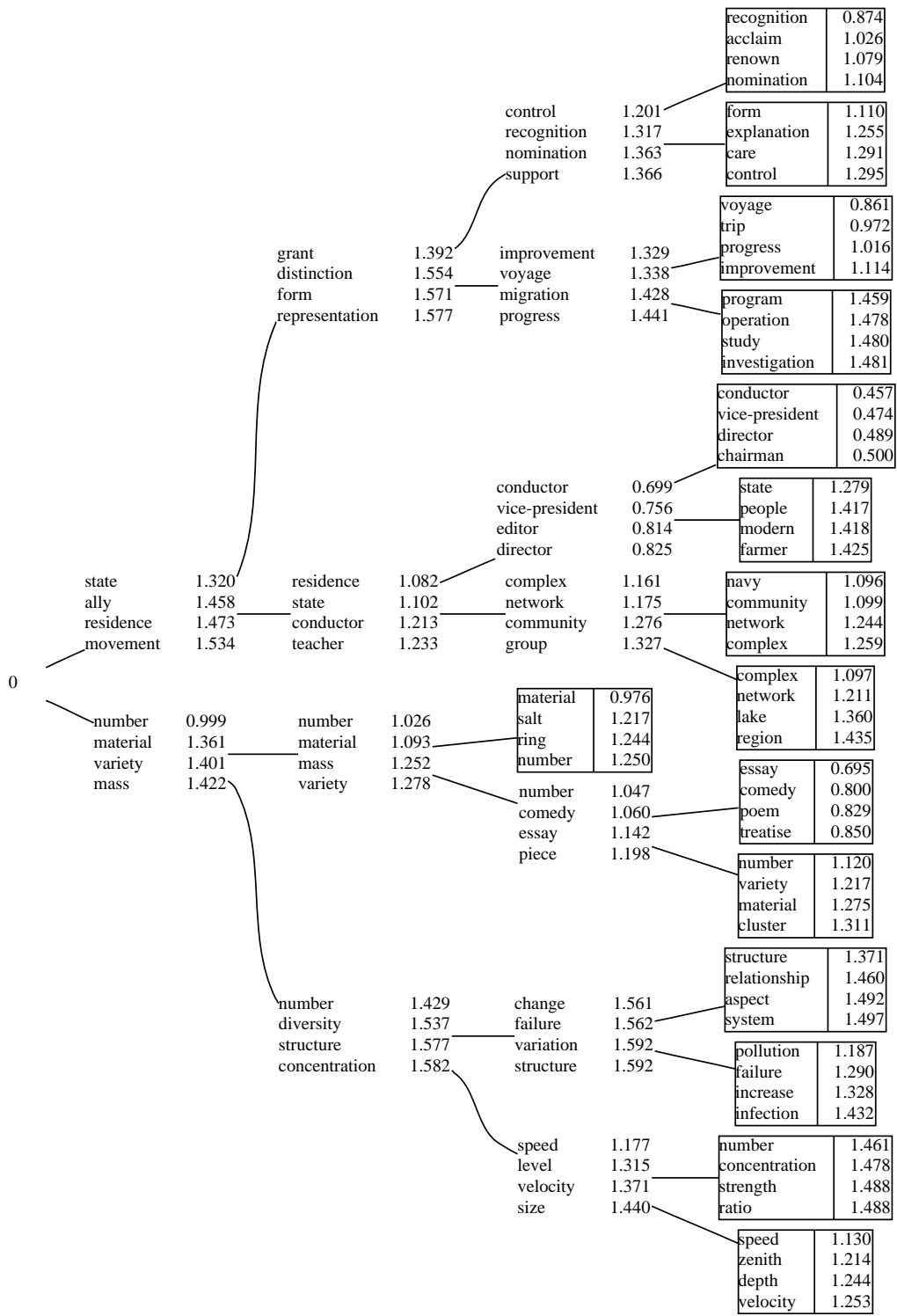}}}
\caption{Noun Clusters for Grolier's Encyclopedia}
\label{grol}
\end{figure*}

Figure \ref{grol} shows the four closest nouns to the centroid of each
of a set of hierarchical clusters derived from verb-object pairs
involving the 1000 most frequent nouns in the June 1991 electronic
version of Grolier's Encyclopedia (10 million words).

\section{MODEL EVALUATION}
The preceding qualitative discussion provides some indication of what
aspects of distributional relationships may be discovered by
clustering. However, we also need to evaluate clustering more
rigorously as a basis for models of distributional relationships. So,
far, we have looked at two kinds of measurements of model quality: (i)
relative entropy between held-out data and the asymmetric model, and
(ii) performance on the task of deciding which of two verbs is more
likely to take a given noun as direct object when the data relating
one of the verbs to the noun has been witheld from the training data.

The evaluation described below was performed on the largest data set
we have worked with so far, extracted from 44 million words of 1988
Associated Press newswire with the pattern matching techniques
mentioned earlier. This collection process
yielded 1112041 verb-object pairs. We selected then the subset
involving the 1000 most frequent nouns in the corpus for clustering,
and randomly divided it into a training set of 756721 pairs and a test
set of 81240 pairs.

\subsection{Relative Entropy}

\begin{figure}
\setlength{\epsfxsize}{3in}
\centerline{\mbox{\epsffile{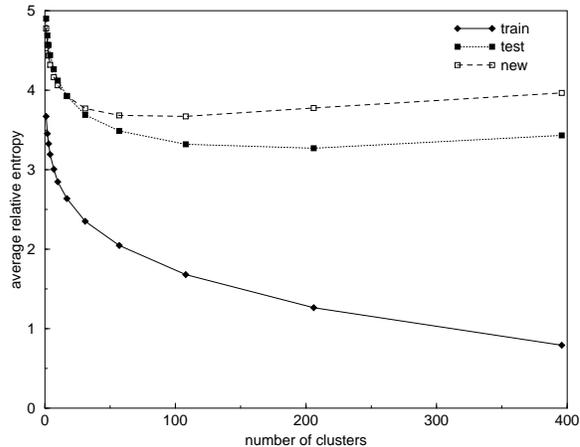}}}
\caption{Asymmetric Model Evaluation, AP88 Verb-Direct Object Pairs}
\label{rel-entropy}
\end{figure}

Figure \ref{rel-entropy} plots the average relative entropy of several
data sets to asymmetric clustered models of different sizes, given by
\[ \sum_n D(t_n||\hat{p}_n) \]
where $t_n$ is the relative frequency distribution of verbs taking $n$
as direct object in the test set. For each critical value of $\beta$, we
show the relative entropy with respect to the asymmetric model
based on $C_\beta$ of the training set (set {\em train}), of
randomly selected held-out test set (set {\em test}), and of held-out
data for a further 1000 nouns that were not clustered (set {\em new}).
Unsurprisingly, the training set
relative entropy decreases monotonically. The test set relative
entropy decreases to a minimum at 206 clusters, and then starts
increasing, suggesting that larger models are overtrained.

The new noun test set is intended to test whether clusters based on
the 1000 most frequent nouns are useful classifiers for the
selectional properties of nouns in general. As the figure shows, the
cluster model provides over one bit of information about the
selectional properties of the new nouns, but the overtraining effect
is even sharper than for the held-out data involving the 1000
clustered nouns.

\subsection{Decision Task}
\begin{figure}
\setlength{\epsfxsize}{3in}
\centerline{\mbox{\epsffile{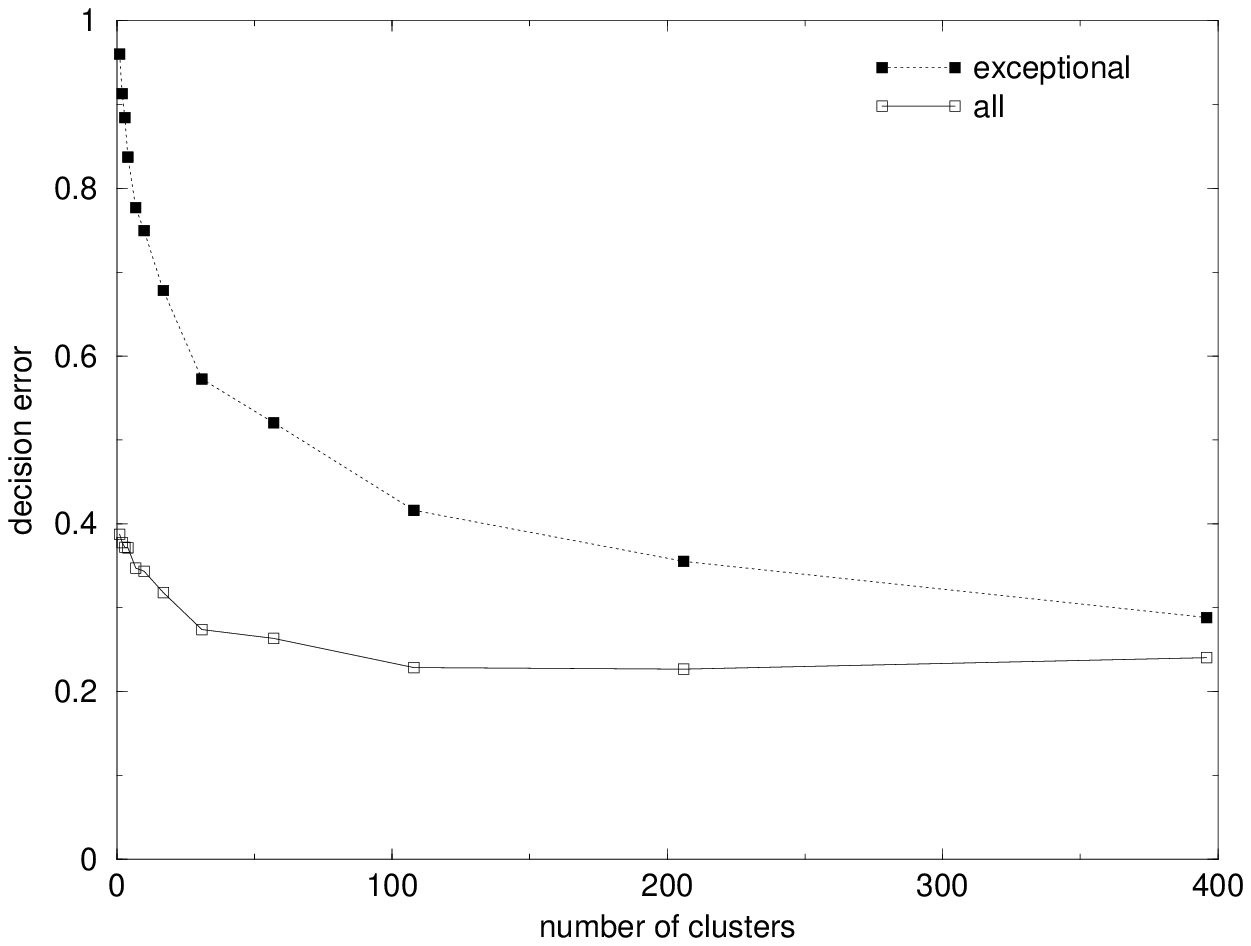}}}
\caption{Pairwise Verb Comparisons, AP88 Verb-Direct Object Pairs}
\label{pairwise}
\end{figure}

We also evaluated asymmetric cluster models on a verb decision task
closer to possible applications to disambiguation in language
analysis. The task consists judging which of two verbs $v$ and $v'$ is
more likely to take a given noun $n$ as object, when all occurrences
of $(v,n)$ in the training set were deliberately deleted. Thus this
test evaluates how well the models reconstruct missing data in the
verb distribution for $n$ from the cluster centroids close to $n$.

The data for this test was built from the training data for the
previous one in the following way, based on a suggestion by
Dagan {\em et al.}  \shortcite{Dagan+al:contextual}. A small
number (104) of $(v,n)$ pairs with a fairly frequent verb (between 500
and 5000 occurrences) was randomly picked, and all occurrences of each
pair in the training set were deleted. The resulting training set was
used to build a sequence of cluster models as before. Each model was
used to decide which of two verbs $v$ and $v'$ are more likely to
appear with a noun $n$ where the $(v,n)$ data was deleted from the
training set, and the decisions compared with the corresponding ones
derived from the original event frequencies in the initial data set.
More specifically, for each deleted pair $(v,n)$ and each verb $v'$
that occurred with $n$ in the initial data either at least twice as
frequently or at most half as frequently as $v$, we compared the
sign of $\log \hat{p}_n(v)/\hat{p}_n(v')$ with that of $\log
p_n(v)/p_n(v')$ for the initial data set. The error rate for each
model is simply the proportion of sign disagreements in the selected
$(v,n,v')$ triples. Figure
\ref{pairwise} shows the error rates for each model for all the
selected $(v,n,v')$ ({\em all}) and for just those {\em exceptional}
triples in which the log frequency ratio of $(n,v)$ and $(n,v')$
differs from the log marginal frequency ratio of $v$ and $v'$. In
other words, the exceptional cases are those in which predictions
based just on the marginal frequencies, which the initial one-cluster
model represents, would be consistently wrong.

Here too we see some overtraining for the largest models considered,
although not for the exceptional verbs.

\section{CONCLUSIONS}
We have demonstrated that a general divisive clustering procedure for
probability distributions can be used to group words according to
their participation in particular grammatical relations with other
words. The resulting clusters are intuitively informative, and can be
used to construct class-based word coocurrence models with substantial
predictive power.

While the clusters derived by the proposed method seem in many cases
semantically significant, this intuition needs to be grounded in a
more rigorous assessment. In addition to predictive power evaluations
of the kind we have already carried out, it might be worth comparing
automatically-derived clusters with human judgements in a suitable
experimental setting.

Moving further in the direction of class-based language models, we
plan to consider additional distributional relations (for instance,
adjective-noun) and apply the results of clustering to the grouping of
lexical associations in lexicalized grammar frameworks such as
stochastic lexicalized tree-adjoining grammars
\cite{Schabes:SLTAG}.

\section{ACKNOWLEDGMENTS} We would like to thank Don Hindle for making
available the 1988 Associated Press verb-object data set, the Fidditch
parser and a verb-object structure filter, Mats Rooth for selecting
the objects of ``fire'' data set and many discussions, David Yarowsky
for help with his stemming and concordancing tools, and Ido Dagan for
suggesting ways of testing cluster models.

\end{document}